\documentclass[letterpaper, 10 pt, conference]{ieeeconf}  

\IEEEoverridecommandlockouts                              

\usepackage{amsmath}
\usepackage{esvect}
\usepackage{mathtools}
\usepackage{amssymb}
\usepackage{amsfonts}
\usepackage{mathabx}

\usepackage[algo2e]{algorithm2e} 
\usepackage{algorithm}
\usepackage{algorithmic}

\usepackage{booktabs}

\usepackage{cite}

\newtheorem{assumption}{Assumption}

\title{\LARGE \bf
Model-Free Large-Scale Cloth Spreading With Mobile Manipulation: Initial Feasibility Study
}

\author{Xiangyu Chu$^{1,2,\dagger}$, Shengzhi Wang$^{1,\dagger}$, Minjian Feng$^{2}$, Jiaxi Zheng$^{3}$, Yuxuan Zhao$^{4}$, \\Jing Huang$^{2}$, and K. W. Samuel Au$^{1,2}$
\thanks{This work was supported in part by Multiscale Medical Robotics Centre, InnoHK, in part by Chow Yuk Ho Technology Centre of Innovative Medicine, The Chinese University of Hong Kong, and in part by the Research Grants Council (RGC) of
Hong Kong under Grant 14211320.}
\thanks{$^{1}$Department of Mechanical and Automation Engineering, The Chinese University of Hong Kong, Hong Kong SAR; $^{2}$Multi-Scale Medical Robotics Center, Hong Kong SAR; $^{3}$School of Engineering, Westlake
University, China; $^{4}$Department of Mechanical and Aerospace Engineering, The Hong Kong University of Science and Technology, Hong Kong SAR}%
}

\begin{document}

\maketitle
\thispagestyle{empty}
\pagestyle{empty}

\begin{abstract}
Cloth manipulation is common in domestic and service tasks, and most studies use fixed-base manipulators to manipulate objects whose sizes are relatively small with respect to the manipulators' workspace, such as towels, shirts, and rags. In contrast, manipulation of large-scale cloth, such as bed making and tablecloth spreading, poses additional challenges of reachability and manipulation control. To address them, this paper presents a novel framework to spread large-scale cloth, with a single-arm mobile manipulator that can solve the reachability issue, for an initial feasibility study. On the manipulation control side, without modeling highly deformable cloth, a vision-based manipulation control scheme is applied and based on an online-update Jacobian matrix mapping from selected feature points to the end-effector motion. To coordinate the control of the manipulator and mobile platform, Behavior Trees (BTs) are used because of their modularity. Finally, experiments are conducted, including validation of the model-free manipulation control for cloth spreading in different conditions and the large-scale cloth spreading framework. The experimental results demonstrate the large-scale cloth spreading task feasibility with a single-arm mobile manipulator and the model-free deformation controller.









\end{abstract}

\section{INTRODUCTION}
Cloth manipulation is critical because it is basic for many downstream applications such as household service \cite{MillerLaundry2012}, elderly care \cite{PuthuveetilUncovered2022}, and surgery scenarios \cite{ThananjeyanSurgical2017}. Cloth manipulation tasks are challenging since the cloth is highly deformable and difficult to be modeled accurately. In this work, we consider one of the common cloth manipulation tasks, spreading \textit{large-scale cloth}\footnote{For large-scale cloth, its size is larger than the workspace of the manipulator's end-effector. This means that a fixed-base manipulator may not be able to finish cloth manipulation tasks such as a spreading task.} over a supporting surface, by using model-free manipulation control and a single-arm mobile manipulator.

\begin{figure}[h]
\includegraphics[width=2.8 in]{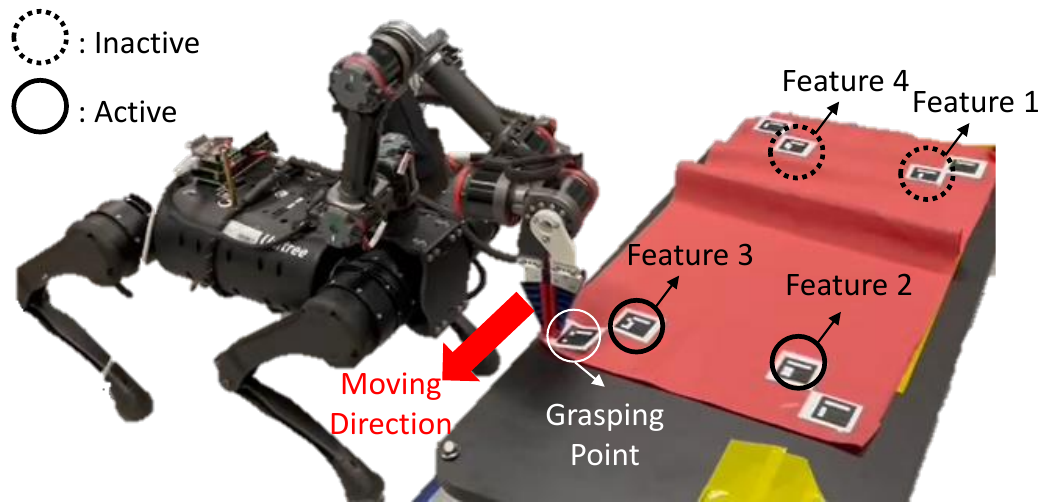}
\caption{A single-arm mobile manipulator is spreading cloth with Grasping Point 3. The end-effector cannot reach Grasping Point 1 if the mobile platform does not move. To reduce unnecessary movement, it is also expected to manipulate the features as much as possible toward the target in one standing position. For example, instead of manipulating corners (features) one by one, Features 2 and 3 can be manipulated to the target (white circles) together. Overall, for large-scale cloth spreading, a mobile platform is necessary to be included and the manipulation control should be able to cover multiple features if possible.  All features and grasping points are selected around the corners to focus on the feasibility study.}
\label{disclosing_two_issues}
\end{figure}

The research on cloth manipulation has a long history, ranging from grasping point selection \cite{ YAMAZAKIGrasp2014, GoldbergGrasp2019, QianGrasp2020} to many specific tasks such as folding \cite{AbbeelGrasp2010}, unfolding \cite{WillimonUnfolding2011}, smoothing \cite{SharmaSmoothing2022}, hanging \cite{DoumanoglouHanging2014}, and even dressing \cite{KragicDress2020}. The studied objects are usually \textit{small-scale} such as towels, shirts, and rags, where fixed-based manipulator(s) can provide sufficient workspace for manipulation. A common paradigm for their manipulation is to first sense the object, then select the grasp point(s), and finally perform control actions to complete tasks (iteratively) \cite{SanchezSurvey2018}. 
Except for the small-scale cloth-like objects, the manipulation of large-scale objects like bed sheets and tablecloths is becoming appealing. 

Among large-scale cloth manipulation applications, a typical one is bed making, where a bed sheet is expected to cover a bed. \cite{GoldbergBed2017} investigated the bed-making task by using deep imitation learning to choose the grasp point. To reach both sides of the bed, a mobile manipulator was used. However, the sheet's bottom side was fixed to the base of the bed frame, which heavily simplified the deformation object manipulation step. Specifically, once the manipulator grasped the sheet, it would directly stretch the sheet towards the top of the bed without the need to consider the effect on the sheet's bottom side.
In another example of manipulating large-scale objects, tablecloth spreading, two mobile manipulators worked together to grasp the cloth at two grasping points and
then spread it on a table \cite{KragicDress2020}. Although multiple mobile manipulators can make large-scale cloth manipulation easier, this approach requires a higher cost and may not be available in a small space. In this paper, we will focus on investigating the feasibility of using a single-arm mobile manipulator for large-scale cloth spreading that generalizes in most large-scale deformable objects. 


To approach more general large-scale cloth spreading, unlike \cite{GoldbergBed2017}, we will not fix any sides of the cloth; in other words, the cloth can move freely on the supporting surface. This case demands more effort to manipulate the cloth given only a single grasping point. For example, pulling the cloth along a fixed direction as in \cite{GoldbergBed2017} may make a part of the large-scale cloth out of the supporting surface, which may fail the task. 
On the other hand, to reduce the mobile platform movement, a controller that can manipulate the area of the cloth as large as possible is preferred, instead of manipulating the corners one by one. 

Due to the challenges of modeling deformable objects accurately, a common practice is to manipulate the vision-based features (see Fig. \ref{disclosing_two_issues}) representing cloth conditions, which can get rid of modeling and simulating deformation behaviors. Specifically, a mapping matrix, also called the Jacobian matrix, is derived to describe the relationship between the end effector motion and the feature change induced by the end effector. However, such a Jacobian matrix works locally \cite{BerensonDeformation2013}. Thus, the Jacobian matrix needs to be updated online after it has been estimated offline. \cite{DavidDeformation2013} was the first time to apply \textit{Broyden update rule} to estimate the deformation Jacobian matrix of elastic bodies iteratively. Later on, this update rule has been widely used in shape control \cite{DavidDeformation2018}. To fold the cloth, \cite{CuiDeformable2020} designed two-stage Jacobian update rules. The first stage (rigidity stage) is to use the ``approximate rigidity” rule to construct the Jacobian matrix when the error between current feature positions and desired points is large; the second stage (deformation stage) is to use the Broyden method to involve the deformation effect in the Jacobian matrix when the error is small. However, their method was only implemented in a simulator and no physical experiments were provided. Spreading cloth has an opposite sequence. In this work, we mainly consider the deformation stage and apply the vision-based manipulation control with an online-update Jacobian matrix. 

Besides the manipulation control, other components are indispensable such as mobile platform control, to demonstrate the overall feasibility. The introduction of the mobile platform complicates the workflow and more sub-systems are involved to perform complex behaviors and conduct multiple objectives together. Previously, people used Finite State Machines (FSMs) to coordinate multiple sub-systems and goals, but the state transition in FSMs is one-way and does not perform well on modularity \cite{FSM}. Behavior Trees (BTs) are alternative solutions because of their modularity, re-usability, and reactivity \cite{DomínguezBT2022}. \cite{BagnellManipulation2012} is the first work to apply BT to robotic manipulation, where several components, including perception, planning, and control were integrated under a BT-based logical structure. Later on, the BT method was applied to mobile manipulation which can improve manipulation capability. For example, \cite{RovidaBT2017} designed an extended BT, that integrates coherently scripted and planned procedures, to achieve a flexible programming paradigm. Their method was applied in kitting tasks where a mobile manipulator is required to navigate to various containers and then pick and place objects. Following the same paradigm, in this work, we will use BTs to coordinate the legged robot control and manipulator control. Moreover, BT's benefits allow us to add more modules easily in future development.

The main contributions of this work are\\
1) Proposing a model-free large-scale cloth spreading framework based on behavior trees that allows us to manipulate cloth with a mobile manipulator;\\
2) Demonstrating the effectiveness of the control algorithm and proposed framework experimentally, thus validating the feasibility of using mobile manipulation for large-scale model-free cloth spreading.




\section{PROBLEM STATEMENT}
Given a piece of cloth and a flat supporting surface (e.g., a table), we consider the task of manipulating the cloth from an initial condition to lay it flat on the surface by using a single-arm mobile manipulator; at the same time, the collision between the supporting platform and the manipulator should be avoided.

\begin{assumption}
The contact between the cloth and the end-effector is fixed, referred to as fixed-point contact in our previous work \cite{HuangDOM2021}.
\end{assumption}
\begin{assumption}
All selected features are always visible and measurable.
\label{visible_measurable}
\end{assumption}
\begin{assumption}
For an initial feasibility study, we only consider the cloth condition that will not result in local minima of the manipulation control because the manipulation control problem is highly underactuated from a control perspective and simple Inverse Kinematics (IK) control is easily subject to local minima \cite{Chu2022}. Although other methods like motion planning and grasping point selection can help avoid the local minima, this aspect is out of the scope of this paper.
\end{assumption}

\section{LARGE-SCALE CLOTH SPREADING FRAMEWORK}

\begin{figure}[t!]
            \includegraphics[width=0.48\textwidth, scale=1]{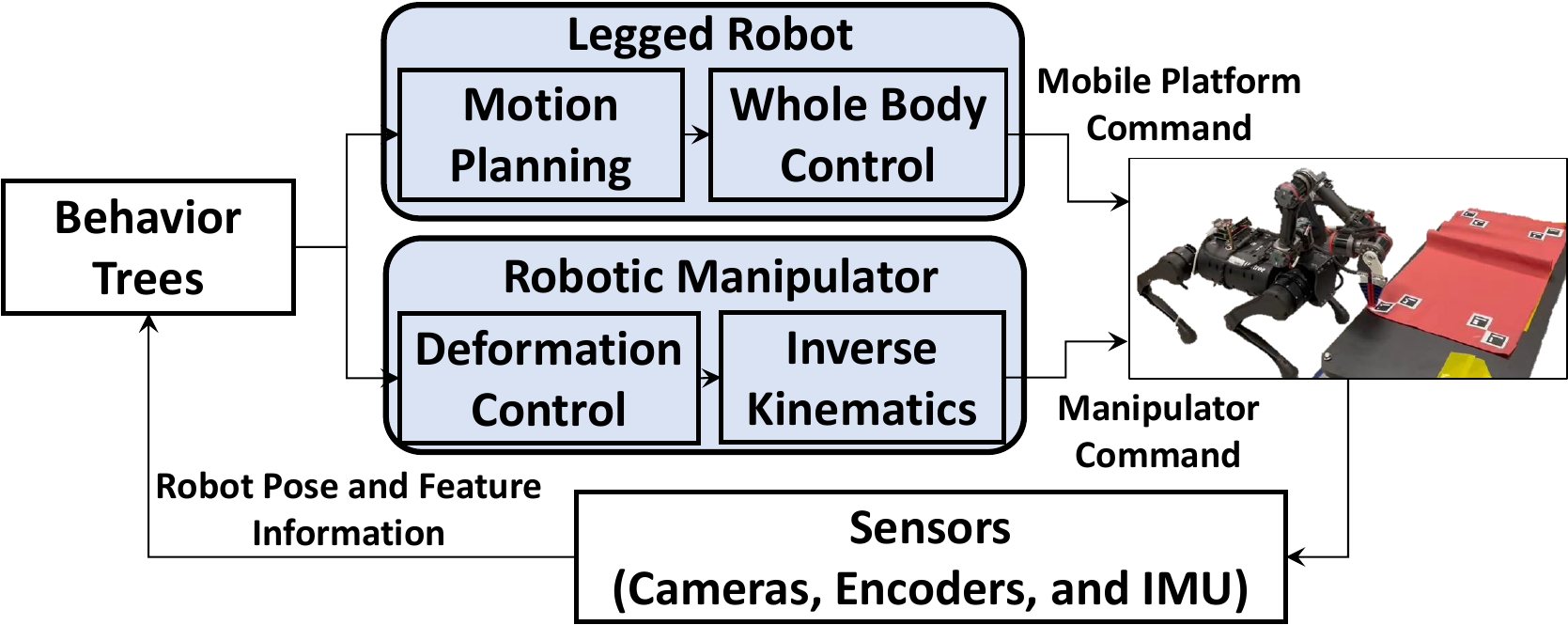}
            \caption{Schematics of the proposed model-free large-scale cloth spreading framework.
            }
            \label{framework}
\end{figure}

\begin{figure}[hbpt!]
            \includegraphics[width=0.45\textwidth, scale=1]{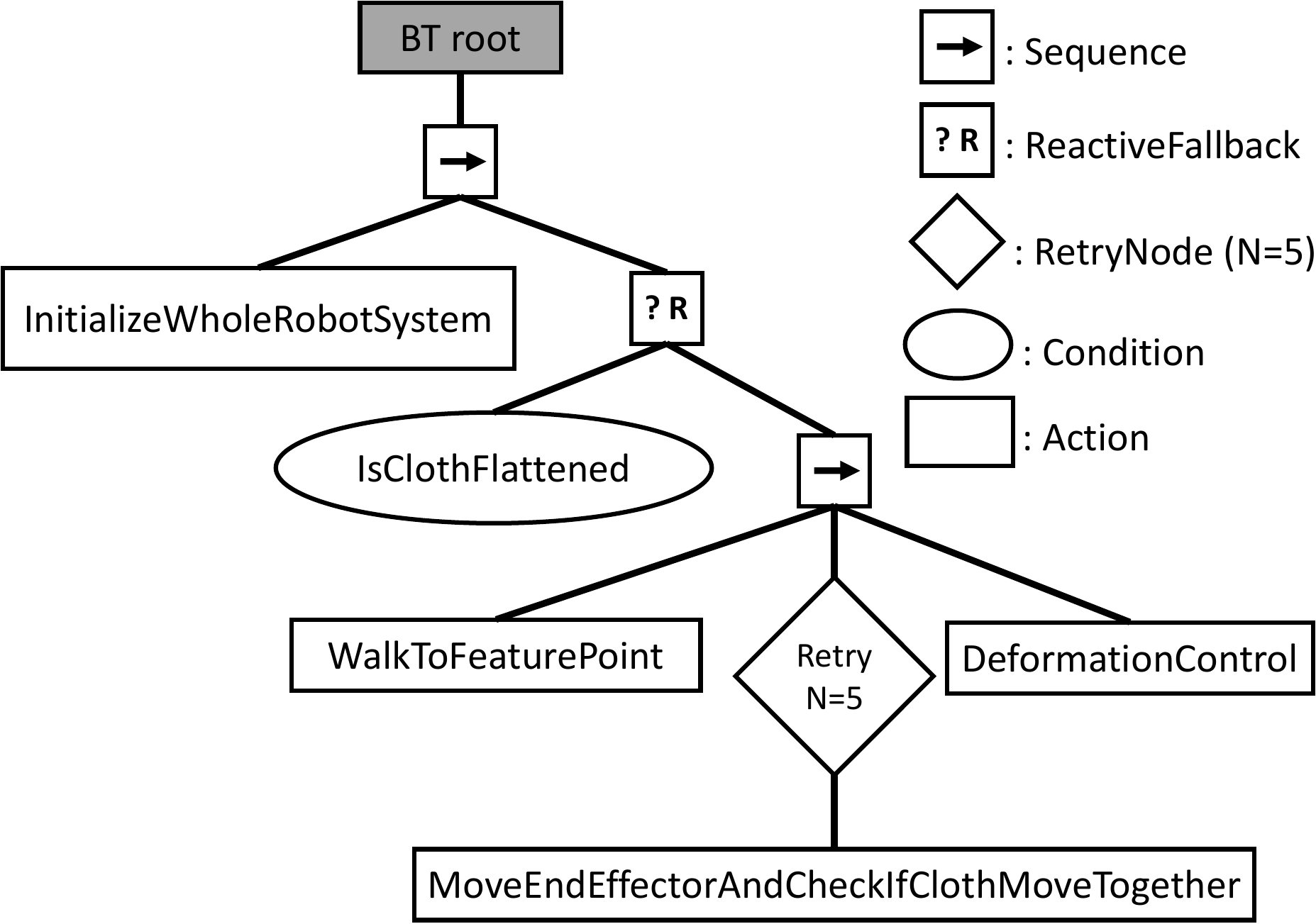}
            \caption{Behavior tree of the proposed model-free large-scale cloth spreading framework.
            }
            \label{fig:BT}
\end{figure}

\subsection{Overview}
We propose a BT-based framework for cloth spreading, as shown in Fig. \ref{framework}, which consists of two main parts: a mobile platform and a robotic manipulator.  Two parts are coordinated by BTs. One is the legged robot (Unitree A1), allowing us to reach any feasible space. Another is the six Degrees of Freedom (DoFs) robotic arm (INNFOS GLUON). A two-finger gripper is attached to the arm. This framework is also applicable to other mobile platforms such as wheeled robots and legged-wheeled robots.

We decouple the manipulator control from the mobile platform control. The floating-base manipulator is controlled by an IK controller. Its end effector's command is from the deformation controller. The disturbance from the manipulator can be handled by the mobile platform control. The trajectory of the mobile platform is planned by an MPC-based trajectory optimization that is solved by OCS2 \cite{OCS2}. The planned trajectory is tracked by a whole body control scheme \cite{qiayuanliao}. The MPC-based planner calculates feedforward trajectories at the update rate of 100 Hz and with a time horizon of 1 s. The mobile platform's state feedback is updated at 500 Hz and obtained from a state estimator that fuses measurements from cameras, encoders, and IMU. The whole body controller of the mobile platform runs at 500 Hz. The feature position is updated at the rate of 30 Hz.

\subsection{Behavior Trees (BTs)}
Compared with one-way FSMs, BTs can alleviate the limitations on modularity, re-usability, and reactivity. Following the terms in \cite{DomínguezBT2022}, a standard BT has one root node, and the execution starts from this node. Signals (ticks) are then sent to a leaf node. After the leaf node executes, the status is sent to its parent. This mechanism makes BTs different from FSMs.

Fig. \ref{fig:BT} shows the BT design for the proposed model-free large-scale cloth spreading framework. Several actions are defined as follows:\\
1) \textit{InitializeWholeRobotSystem:} This action controls the manipulator to its home position in the joint space, and the mobile platform to stand up.\\
2) \textit{WalkToFeaturePoint:} This action moves the mobile platform to the feature point that is nearest to the robot but has not yet reached its target
.\\
3) \textit{MoveEndEffectorAndCheckIfClothMoveTogether:} Due to the noise during feature detection and the resulting inaccurate position measurement along $z$ axis, the end-effector cannot grab the grasping point occasionally. To ensure the grasping point is grabbed, we develop this skill to move the end-effector horizontally in Cartesian space after the end-effector attempts to grab the grasping point, and then check whether its closest feature point is also moved in a constant period. If the grasping point was not grabbed, we add a uniformly distributed offset to the vertical direction of the desired position of the end-effector and retry the checking process until the cloth is grabbed. \\
4) \textit{DeformationControl:} Once the contact is established, this action executes the deformation control, as shown in \ref{subsec::deformationcontrol}. 

In summary, BT initializes the whole robot system and subsequently executes the large-scale cloth spreading task. During the task, BT constantly checks whether the cloth is flattened or not. When the cloth is not flattened, BT specifies the grasping point. The mobile platform's trajectory is generated by the motion planner and then tracked by the whole body controller (see Fig. \ref{framework}).
For the contact establishment attempt, five chances are given by \textit{RetryNode}. The \textit{DeformationControl} action is only triggered after a successful attempt. However, if no contact is established after five times, the \textit{Sequence} node restarts from the \textit{WalkToFeaturePoint} action node and then retries the contact establishment. This process repeats until the cloth is grabbed. In the end, the whole robot system stops until the cloth is flattened. 


\section{MODEL-FREE DEFORMATION CONTROL IN CLOTH SPREADING}
\subsection{Mobile Manipulator}
Considering a single-arm legged manipulator, we can model it as a manipulator mounted on a floating base, and its joint displacement vector is
\begin{equation}
    \begin{aligned}
        \boldsymbol{q}=[p_x, p_y, p_z, r_x, r_y, r_z, \theta_1, ..., \theta_N]^T,
    \end{aligned}
\end{equation}
which consists of the six-DoF base pose of the legged robot and $N$-DoF arm joint positions. The manipulator's end-effector position $\boldsymbol{x} \in \mathbb{R}^p$ is described by $\boldsymbol{x} = \boldsymbol{f}(\boldsymbol{q})$, and its differential kinematic equation is given by 
\begin{equation}
    \begin{aligned}
        \dot{\boldsymbol{x}} = \frac{\partial \boldsymbol{f}}{\partial \boldsymbol{q}} \dot{\boldsymbol{q}}.
        \label{differential_kinematic}
    \end{aligned}
\end{equation}

In this paper, we consider the mobile manipulator control as a kinematic control mode while manipulating the cloth. Once obtaining the command of the end-effector, the mobile manipulator can conduct the command based on the inverse kinematics by inverting (\ref{differential_kinematic}).

\subsection{Deformation Description}
Following \cite{DavidDeformation2013}, we describe the deformation in a model-free paradigm. Specifically, to manipulate the deformable cloth, $k$ feature points are selected on the cloth. The $i$-th feature point is denoted as
\begin{equation}
    \begin{aligned}
        \boldsymbol{s}_i = [x_i, y_i, z_i]^T,
        \label{feature_point_def}
    \end{aligned}
\end{equation}
and $\boldsymbol{s}_i$ is measurable according to Assumption \ref{visible_measurable}. For a compact notation, the feature point vector can be denoted as
\begin{equation}
    \begin{aligned}
        \boldsymbol{s} = [\boldsymbol{s}_1^T, \boldsymbol{s}_2^T, ..., \boldsymbol{s}_k^T]^T \in \mathbb{R}^{3k}.
        \label{feature_point_def}
    \end{aligned}
\end{equation}
\begin{assumption}
    When manipulating the deformable cloth, we assume each feature point can be locally described by a smooth function, $\boldsymbol{s}_i = \boldsymbol{g}_i (\boldsymbol{x})$. Then, the relationship between the feature point vector and the end-effector position is $\boldsymbol{s} = \boldsymbol{g} (\boldsymbol{x})$, where $\boldsymbol{g} = [\boldsymbol{g}_1^T, ..., \boldsymbol{g}_k^T]^T$. Note that $\boldsymbol{g}$ is unknown.
\end{assumption}

Based on these feature points, a $m$-DoF deformation task $\boldsymbol{y}$ can be defined as $\boldsymbol{y} = \boldsymbol{r}(\boldsymbol{s})$. The task can be designed explicitly, as mentioned in \cite{DavidDeformation2013}. To implement the task, a velocity controller of the end-effector can be designed as
\begin{equation}
    \begin{aligned}
        \boldsymbol{u} = \boldsymbol{J}^\dagger \boldsymbol{K} (\boldsymbol{y}^* - \boldsymbol{y}), ~~ \boldsymbol{J}(\boldsymbol{x}) := \frac{\partial \boldsymbol{r}(\boldsymbol{s})}{\partial \boldsymbol{s}} \frac{\partial \boldsymbol{g}(\boldsymbol{x})}{\partial \boldsymbol{x}},
        \label{velocity_controller}
    \end{aligned}
\end{equation}
where $\boldsymbol{K} \succ 0$ and $(\cdot)^\dagger$ denotes a left pseudoinverse operation if $m>p$ or a right pseudoinverse operation if $m<p$. $\boldsymbol{J}(\boldsymbol{x})$ is called \textit{Deformation Jacobian Matrix}, and it maps the end-effector motion to the evolution of the defined task.

\subsection{Deformation Control}
\label{subsec::deformationcontrol}
Since the task is user-defined, $\frac{\partial \boldsymbol{r}}{\partial \boldsymbol{s}}$ is known. However, $\frac{\partial \boldsymbol{g}}{\partial \boldsymbol{x}}$ is unknown, thus the deformation Jacobian matrix cannot be expressed explicitly. To achieve the task, we need to estimate the Jacobian matrix and an online Broyden update rule will be used, which can calculate an estimated $\hat{\boldsymbol{J}}$ at each time step. Specifically, the update rule is 
\begin{equation}
    \begin{aligned}
        & \hat{\boldsymbol{J}}({j+1}) \\ & = \hat{\boldsymbol{J}}({j}) + \alpha \frac{\Delta \boldsymbol{y}({j+1}) - \hat{\boldsymbol{J}}(j) \Delta \boldsymbol{x}(j+1)}{{\Delta \boldsymbol{x}(j+1)}^T \Delta \boldsymbol{x}(j+1)} {\Delta \boldsymbol{x}(j+1)}^T,
        \label{Broyden_update_rule}
    \end{aligned}
\end{equation}
where $(\cdot)({j})$ denotes the variable at the $j$-th step, 
$\Delta \boldsymbol{x}(j+1) \neq \boldsymbol{0}$, $\Delta \boldsymbol{y} (j+1) = \boldsymbol{y}({j+1}) - \boldsymbol{y}({j})$, $\Delta \boldsymbol{x}(j+1) = \boldsymbol{x}({j+1}) - \boldsymbol{x}(j)$, and $\alpha \in (0, 1]$ is a scalar weighting between the update rate and accuracy. Such an update rule works for \textit{slow} manipulation process and it does not need an analytical model of the deformable object. Therefore, the velocity controller of the end-effector is updated to
\begin{equation}
    \begin{aligned}
        \boldsymbol{u}(j+1) = \boldsymbol{\hat{J}}^\dagger (j+1) \boldsymbol{K} (\boldsymbol{y}^* - \boldsymbol{y}(j+1)).
        \label{updated_velocity_controller}
    \end{aligned}
\end{equation}

\section{Initial Experimental Results}

\subsection{Experimental Setup}
To study the task feasibility, without loss of generality, we use a legged manipulator to manipulate a piece of cloth (72 cm $\times$ 35 cm) within a horizontal plane. We assume the cloth's shape is rectangle-like, but our method is not limited by the shape. The cloth was initially put on a table with a height of 0.28 m. As shown in Fig. \ref{disclosing_two_issues}, four feature points (ArUco markers) close to each corner were selected. An overhead camera (Intel® RealSense™ depth camera D435) was used to detect and locate each feature point, and the depth information was ignored here. A motion capture system (Vicon) was used to locate the legged robot and the overhead camera. In the validation of the deformation controller, all positions were expressed in the body frame of the legged robot, while all were expressed in the Vicon frame in the validation of the framework. 

\begin{figure}[t!]
        \centering
            \includegraphics[width=0.4\textwidth, scale=1]{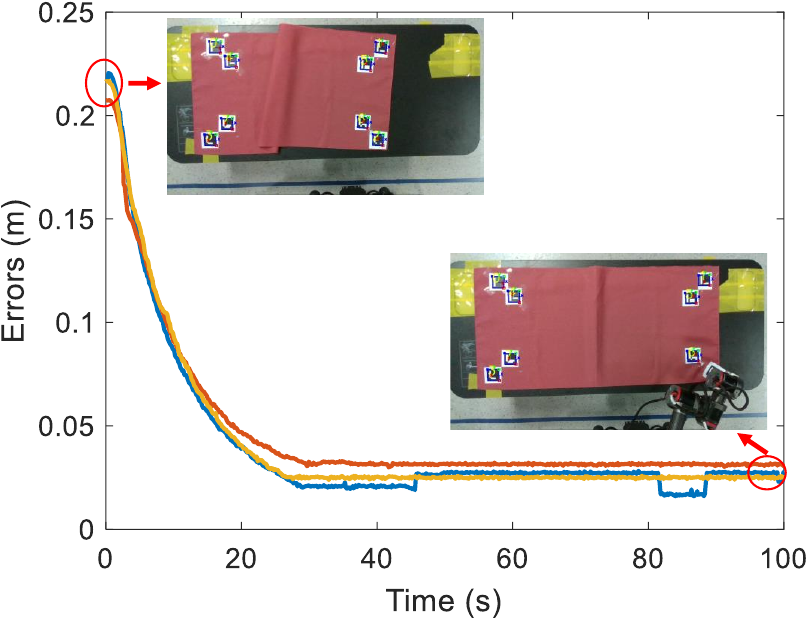}
            \caption{Errors of feature position in cloth deformation condition 1 (3 trials).
            }
            \label{exp_deformation_control config 1}
\end{figure}

\begin{figure}[t!]
        \centering
            \includegraphics[width=0.39\textwidth, scale=1]{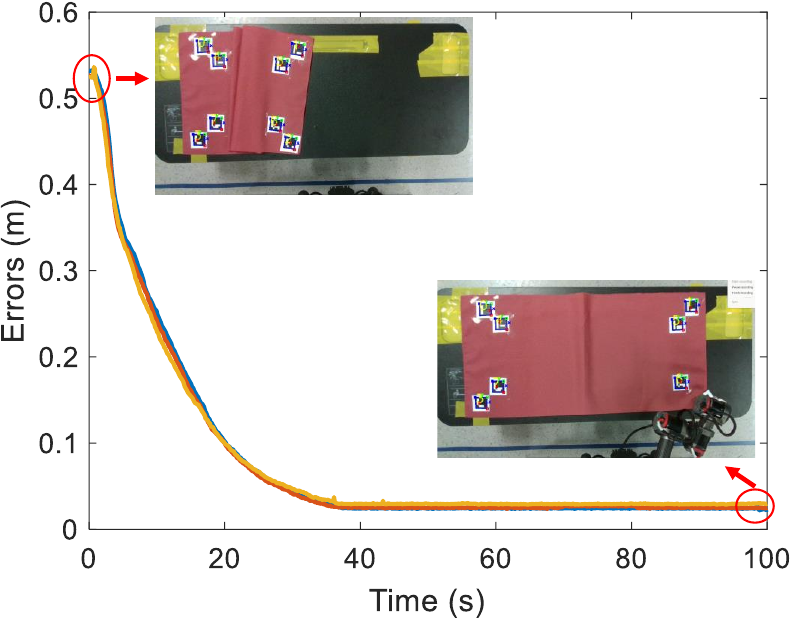}
            \caption{Errors of feature position in cloth deformation condition 2 (3 trials).
            }
            \label{exp_deformation_control config 2}
\end{figure}

\begin{figure*}[t!]
        \centering
            \includegraphics[width=0.8\textwidth, scale=1]{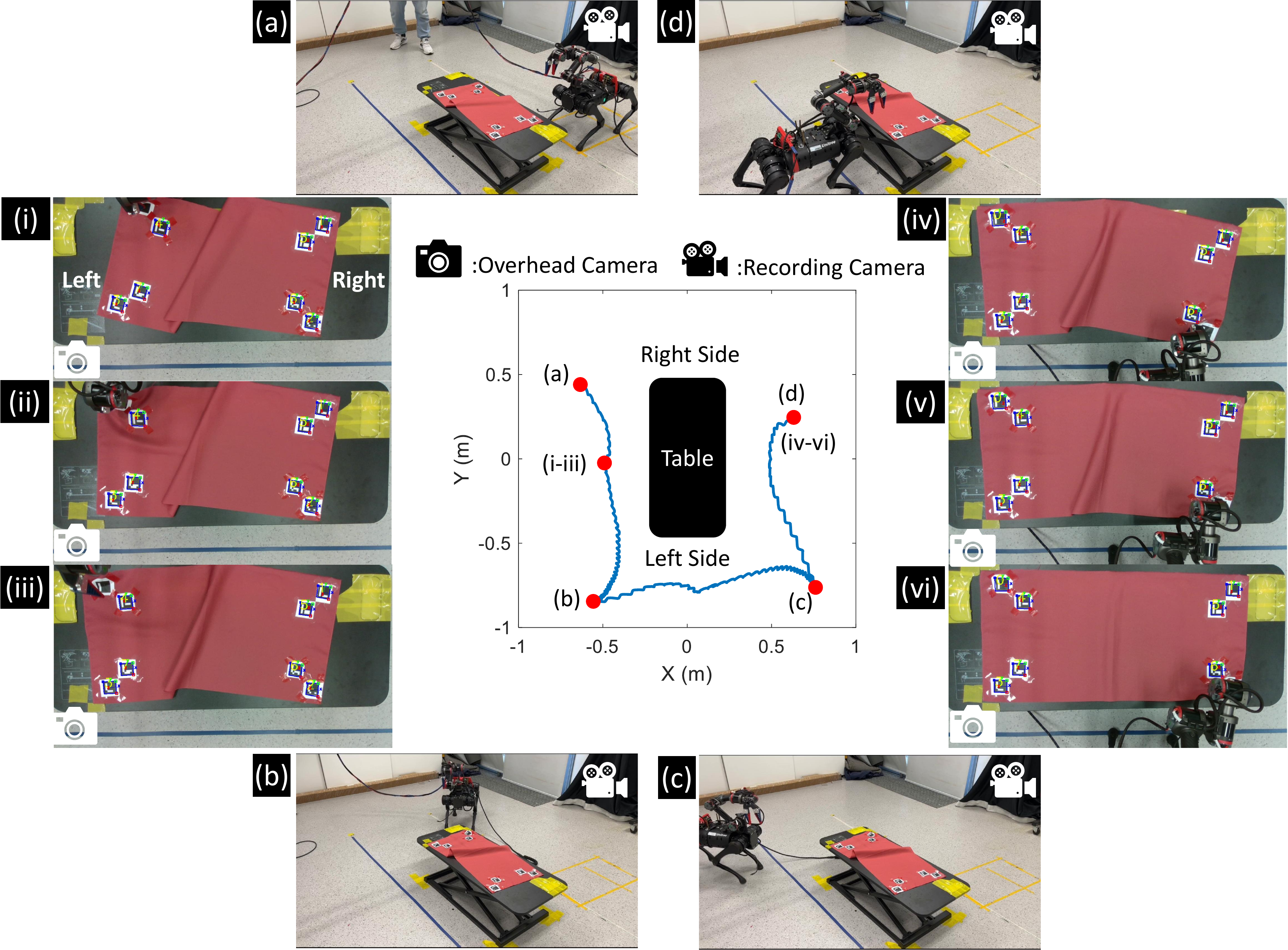}
            \caption{Experiment of a large-scale cloth spreading task with the proposed framework.
            }
            \label{exp_framework}
\end{figure*}

\subsection{Validation of Model-Free Deformation Control}
To verify the effectiveness of the model-free deformation controller (\ref{updated_velocity_controller}), the legged robot was in static and only the manipulator was used. The Jacobian matrix between each feature and the end-effector was initialized to be an identity matrix. The update rate $\alpha$ was set to 0.1. The controller gain matrix was set to $3.5 \boldsymbol{I}_{8 \times 8}$.  To quantify the control performance, an error is defined as $||\boldsymbol{s}_{dsr} - \boldsymbol{s}||$, where $\boldsymbol{s}_{dsr} \in \mathbb{R}^8$ is the desired position for four features and can be predefined. $\boldsymbol{s}_i\in \mathbb{R}^2$ only includes $x$ and $y$ elements where $i = 1,...,4$. Due to diminishing rigidity, the features farther from the grasping point have smaller displacements; in other words, the end-effector motion cannot affect these features significantly. Therefore, we validate the control in the case of regulating Feature 2 and 3 to the target while keeping Feature 1 and 4 static (see Fig. \ref{disclosing_two_issues} for feature distribution).

To this end, two cloth deformation conditions were shown. In the first condition (Fig. \ref{exp_deformation_control config 1}), three trials are displayed and all of them converged to the error of less than 0.032 m after around 28 s. The final cloth condition is almost fully flattened. Fig. \ref{exp_deformation_control config 2} shows another initial deformation condition where the initial positions of Feature 2 and 3 were much farther from the target (initial error: 0.524 m). The model-free deformation controller still can make the error converge to the value of less than 0.030 m, but the manipulation duration became longer (36 s). From the results of the two conditions, a steady state error was found for each trial. This was mainly because the visual measurement noise would affect the feature localization and further inject disturbance into the Jacobian update process. To improve the control performance, a more precise and robust feature localization method is needed, which will be our future work.

\subsection{Validation of Large-Scale Cloth Spreading Framework }

To validate the proposed framework, we set the cloth condition that requires multiple operations at different standing positions. Running BTs helps to specify the grasping points and thus the trajectory of the mobile platform can be planned accordingly. As shown in Fig. \ref{exp_framework}, the mobile platform's trajectory is plotted (blue solid line). Fig. \ref{exp_framework}(a) shows its starting status. Once finishing the first manipulation (Figs. \ref{exp_framework}(i-iii)), it moved to the next location (Fig. \ref{exp_framework}(d)) following the predefined waypoints with a safe distance (see Figs. \ref{exp_framework}(b-c)). When manipulating the cloth at the first standing position, the cloth condition change is shown in Figs. \ref{exp_framework}(i-iii)). With the deformation control, the two feature points on the left side were manipulated to the predefined target. Figs. \ref{exp_framework}(iv-vi) shows the cloth manipulation at the second standing position. After establishing the contact, the manipulator moved two feature points on the right side to the target. At the same time, the feature points on the left side were not affected and the cloth was fully spread finally. 

The experimental results demonstrate the effectiveness of the proposed framework and the feasibility of spreading large-scale cloth in the sense of model free. Of course, the framework can be further improved. For example, the distance between the mobile platform and the table can be smaller for saving occupying space, via motion planning in narrow space (e.g., \cite{Liao2023}). Besides, an optimal grasping point selection strategy can be designed to improve cloth manipulation efficiency and minimize the traveling path of the mobile platform.

\section{DISCUSSIONS}

In this work, we are interested in large-scale cloth spreading with two extra challenges compared with small-scale cloth-like object manipulation. The first challenge is the insufficient workspace for common fixed-base manipulators. The second is deformable object manipulation control which is often simplified in previous work (e.g., the bottom side of the sheet is fixed in \cite{GoldbergBed2017}).  Facing the challenges of reachability and manipulation control, we proposed a model-free large-scale cloth spreading framework that coordinates a mobile platform and a manipulator by using BTs. Based on it, a task feasibility study was successfully conducted. 

Our current algorithms cannot be able to spread the highly crumpled and overlapped cloth, because some visual features may not be visible. An alternative solution is to apply multiple grasps first to make all features visible before using deformation control. On the other hand, although visual features are subject to measurement noise, they allow us to impose more task-oriented constraints on the whole cloth's condition; for example, the cloth can be always kept on a supporting surface or within the visibility area during the whole tasks (e.g., bed making and tablecloth spreading) by enforcing state constraints (e.g., integrating control barrier functions \cite{AmesCBF2016} into the deformation controller). In large-scale cloth spreading, these constraints are useful because a part of the cloth may be out of the visibility set (e.g., the hanging part of the cloth cannot be observed) and thus this part would not be controlled properly before achieving cloth spreading.









\section{CONCLUSIONS}
This paper presented a model-free large-scale cloth spreading framework based on BTs. The introduction of the mobile platform eliminates the limitation on the workspace. Without the need of modeling the cloth, a vision-based deformation control was applied to manipulate the cloth. Both the deformation control and cloth spreading framework have been verified experimentally. The results show the feasibility of spreading large-scale cloth using a single-arm mobile manipulator and a model-free control approach, paving the way to manipulating large-scale cloth with more complex conditions (e.g., highly crumpled conditions). In the future, extra components will be integrated into the current framework, such as onboard perception and action planning modules. Different cloth materials will be tested and the framework will be applied to more practical scenarios like in households.






\section*{ACKNOWLEDGMENT}


The authors want to thank Dr. Bohan Yang and Mr. Yiang Lu for their discussions.

\end{document}